\documentclass[11pt]{article}
\usepackage[final]{acl}
\usepackage{times}
\usepackage{latexsym}
\usepackage[T1]{fontenc}
\usepackage[utf8]{inputenc}
\usepackage{microtype}
\usepackage{inconsolata}
\usepackage{graphicx}
\usepackage{booktabs}
\usepackage{xurl}
\usepackage{array}
\usepackage{xcolor}
\usepackage{tikz}
\usetikzlibrary{positioning,arrows.meta,calc}
\definecolor{hqblue}{HTML}{08519C}
\definecolor{aqorange}{HTML}{D94801}
\newcommand{\hg}{H\textsubscript{G}}
\newcommand{\hq}{H\textsubscript{Q}}
\newcommand{\ag}{AI\textsubscript{G}}
\newcommand{\aq}{AI\textsubscript{Q}}

\title{Emotions and Appraisals in Human and AI-Generated\\Song Lyrics about Queer Lives}
\author{Roman Klinger \and Sabine Weber\\
  Fundamentals of Natural Language Processing, University of Bamberg, Germany\\
  \texttt{\{roman.klinger,sabine.weber\}@uni-bamberg.de}}

\begin{document}
\maketitle

\begin{abstract}
  Songs influence how listeners think about people. With AI music
  services, which write and sing a song about any group on request,
  producing such songs on purpose and at scale has become easy. We
  therefore ask how AI-generated songs represent queer lives, compared
  to human-written songs.  We compare four groups of song lyrics:
  human-composed songs associated with queer communities and other
  human-composed songs from the Genius data set, and AI-generated
  songs for prompts with and without a queer identity from the Sounds
  Queer data set. For the analysis, we apply an emotion model from the
  BamNLP Emotion Hub to predict emotion categories, valence and
  arousal, and appraisals, i.e., how an event mentioned in the text is
  evaluated. Our main findings are as follows: Human queer songs are
  more joyful than other human songs. AI songs are much more joyful
  than human songs in every genre, with or without a queer identity in
  the prompt. What distinguishes the AI's queer songs are the emotion
  trajectories: they move from sadness and confinement to joy and
  self-determination, while human songs show no unified emotional
  story arc.
\end{abstract}

\section{Introduction}

Music shapes how people perceive others. Song lyrics change thoughts
and behaviour toward the groups they depict: sexually aggressive
lyrics increase aggression toward women \citep{fischer2006}, lyrics in
favour of integration reduce prejudice against immigrants
\citep{greitemeyer2014}, and shared songs can soften the boundaries
between conflicting groups \citep{bodner2009}. Media representations
of gay people reduce prejudice when they let audiences feel acquainted
with them \citep{schiappa2005}. How a group appears in music is
therefore part of how a society sees that group. This matters
particularly for queer people, who face prejudice and the stress that
comes with it \citep{meyer2003}: in a survey of more than 100,000
LGBTIQ people in Europe, 55\% had experienced hate-motivated
harassment in the year before, and 54\% avoid holding hands with a
same-sex partner in public \citep{fra2024}.

At the same time, AI-generated music enters streaming platforms at a
fast pace. Services such as Suno\footnote{\url{https://suno.com}} and
Mureka\footnote{\url{https://www.mureka.ai}} write and sing a song on
any topic. In April 2026, 44\% of all tracks uploaded to the streaming
service Deezer, almost 75,000 per day, were fully AI-generated
\citep{deezer2026}. What do these services make of queer lives?
Research on language and image models has seen a shift: earlier models
produced openly harmful representations of queer people
\citep{nozza2022,felkner2023winoqueer}; current systems present
marginalised groups positively, but along narrow scripts
\citep{cheng2023marked}. A positive stereotype remains a stereotype
\citep{czopp2015}, and can still harm the stereotyped group.

This paper makes stereotyping in AI-generated songs measurable. We
ask: \emph{How do human-written song lyrics on queer topics differ
  from AI-generated ones in the emotions and appraisals they
  describe?}  If AI songs about queer people are happier than
human-written ones, this may be because human queer songs are happy
and the AI reproduces the pattern, or because AI songs are happier in
general. We therefore compare four groups in a $2\times2$ design:
human and AI songs, each with and without a queer topic. Songs rarely
name their emotions; they describe situations, and listeners infer the
emotions. Appraisal theories in psychology explain how this works: an
emotion follows from how a person evaluates an event with respect to
their goals and needs \citep{smith1985,scherer2001}. Being rejected by
others can cause anger when the others are held responsible, sadness
when nobody can change the situation, and shame when one blames
oneself. We therefore analyse not only emotion categories but also
such appraisal dimensions.

\section{Related Work}

\subsection{Emotion Theories and Appraisals}
Emotion analysis in natural language processing builds on theories
from psychology, which \citet{scarantino2016} groups into a
motivation, a feeling and an evaluative tradition. Theories of basic
emotions belong to the motivation tradition and provide the label sets
of most emotion corpora: \citet{ekman1992} proposes anger, disgust,
fear, joy, sadness and surprise; \citet{plutchik2001} adds trust and
anticipation. The feeling tradition represents emotions in a
continuous space of valence (how pleasant the experience is) and
arousal \citep[how activated the experiencer
is;][]{russell1980,posner2005}.

Appraisal theories, the evaluative tradition, explain emotions by the
cognitive evaluation of an event with respect to its impact on the
experiencer. The set of appraisal dimensions depends on the theory and
the domain. It often includes whether an event is pleasant and
consistent with one's goals, how much effort it requires (high for
anger and fear), who is responsible for it (oneself for pride or
guilt, others for anger) and who can control it.  \citet{smith1985}
show that six such dimensions distinguish 15
emotions. \citet{scherer2001} describes appraisal as a sequence of
checks \citep{scherer2005}.

Only recently, corpora and models have made appraisal dimensions
explicit for text
\citep{hofmann2020,stranisci2022,wegge2022,troiano2023,greschner2026contarga}. Appraisals
are useful for the analysis of songs because they capture emotions
that are expressed only implicitly. The BamNLP Emotion Hub
\citep{klinger2027hub} brings together 48 emotion corpora with
categories, valence and arousal, appraisals and further annotations in
one format, and compares models across
them.\footnote{\url{https://emotionanalysis.de}} Its general model,
trained jointly on the corpora with human annotations, predicts
emotion categories, valence/arousal and appraisals for English and
German texts; we use it for our analysis.

\subsection{Emotions in Song Lyrics}
A song consists of music and lyrics. Music on its own expresses and
evokes emotions, through learned associations and expectations
\citep{juslin2008}.  Lyrics add who feels what about which
situation. The two interact.  Listeners perceive the emotion of a song
mainly through its melody, but lyrics strengthen sad and angry music
and weaken happy and calm music \citep{ali2006}. Lyrics activate
emotion-related brain areas more in sad than in happy music
\citep{brattico2011}.

We analyse only the lyrics. For mood classification, lyrics can
outperform audio features \citep{hu2010}, and lyrics and music
contribute complementary emotion information
\citep{mihalcea2012}. Large-scale studies used word lists to follow
the valence of popular lyrics over decades and found that they became
more negative \citep{dodds2010,dewall2011,brand2019}. Such lexicons
capture emotion words, not the situations a song describes. In the
humanities, emotion analysis has mostly studied literature
\citep{kim2019survey}; we follow \citet{reagan2016} in also
considering how emotions change over the course of a text.

\begin{figure*}[t]
\centering\small
\begin{tikzpicture}[
  font=\fontfamily{phv}\selectfont\small,
  box/.style={draw, align=center, inner sep=3pt, minimum height=7mm},
  data/.style={box, fill=black!4, text width=2.3cm},
  hum/.style={box, fill=hqblue!15, draw=hqblue, text width=3.4cm},
  ai/.style={box, fill=aqorange!15, draw=aqorange, text width=3.4cm},
  proc/.style={box, fill=white, text width=2.0cm},
  ana/.style={box, fill=black!4, text width=4.2cm, align=left},
  arr/.style={-{Latex[length=2mm]}, thick, black!60},
  line/.style={thick, black!60},
  scale=.78, transform shape]
\node[data] (gen) at (0, 0) {Genius\\3.2M songs};
\node[data] (sq) at (0, -2.6) {Sounds Queer\\5,963 songs};
\node[hum] (hg) at (4.1, 0.75) {\hg: human, not queer};
\node[hum] (hq) at (4.1, -0.45) {\hq: human, queer\\(Wikipedia lists)};
\node[ai] (an) at (4.1, -2.05) {\ag: AI, not queer};
\node[ai] (aq) at (4.1, -3.15) {\aq: AI, queer};
\draw[line] (gen.east) -- ++(0.55, 0) coordinate (g);
\foreach \t in {hg, hq} \draw[arr] (g) |- (\t.west);
\draw[line] (sq.east) -- ++(0.55, 0) coordinate (s);
\foreach \t in {an, aq} \draw[arr] (s) |- (\t.west);
\node[proc] (stz) at (8.3, -0.45) {stanzas};
\node[proc] (mod) at (8.3, -1.75) {\textbf{emotion model}};
\coordinate (b) at ($(hg.east) + (0.6, 0)$);
\foreach \t in {hg, hq, an, aq} \draw[line] (\t.east) -- (\t.east -| b);
\draw[line] (b |- hg.east) -- (b |- aq.east);
\draw[arr] (b |- stz.west) -- (stz.west);
\draw[arr] (stz) -- (mod);
\node[proc] (prof) at (11.7, -0.45) {\textbf{profile}\\of a song};
\node[proc] (cour) at (11.7, -2.95) {\textbf{course}\\of a song};
\draw[line] (mod.east) -- ++(0.6, 0) coordinate (m);
\draw[arr] (m) |- (prof.west);
\draw[arr] (m) |- (cour.west);
\node[ana] (cmp) at (16.1, -0.45) {Q1--Q3: How do the four groups differ in emotions and appraisals?};
\node[ana] (nar) at (16.1, -2.95) {Q4: Do the songs follow a narrative, i.e., a typical course of emotions?};
\draw[arr] (prof.east) -- (cmp.west);
\draw[arr] (cour.east) -- (nar.west);
\end{tikzpicture}
\caption{Lyrics of human and AI-generated songs from two databases
  go into an emotion model, stanza by stanza.}
\label{fig:sketch}
\end{figure*}
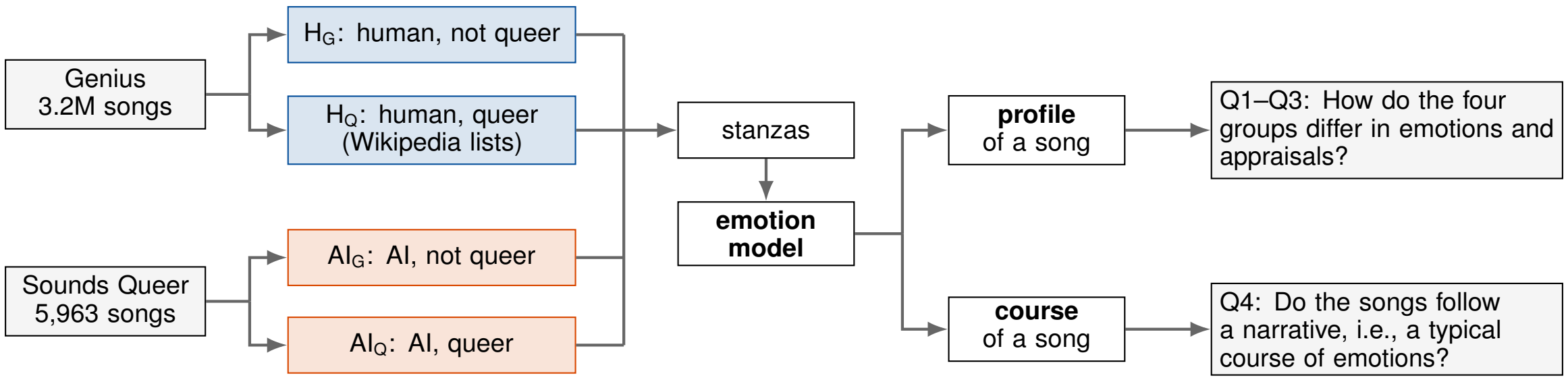

\subsection{Queer Representation in Generative AI}
Benchmarks and analyses show anti-queer bias and the exclusion of
non-binary people in language technology
\citep{nozza2022,felkner2023winoqueer,dev2021}, and
\citet{dhingra2023queer} find stereotypes of sexual identity in
generated text. \citet{cheng2023marked} let models describe personas
of groups marked by race, ethnicity and gender and find seemingly
positive but essentialising words such as \emph{resilient} and
\emph{vibrant}. Text-to-image models reduce non-cisgender identities
to a few visual stereotypes \citep{ungless2023} and amplify
demographic stereotypes
\citep{bianchi2023}. \citet{weber2026soundsqueer} generated songs for
1,200 combinations of genre and identity; in their annotation of 100
songs, 86 are about identity, 43 are love songs, and
self-determination is the most frequent theme (53 songs). We build on
their data, add human songs and analyse emotions and appraisals.

\subsection{Queer Popular Music}
Popular music has long been a space in which queer people find and
build community, often through songs that were not written about them
\citep{whiteley2006,taylor2012}. ``Gay anthems'' are typically
adopted: dance and disco songs, as disco and its dance floors were
central to gay club culture \citep{lawrence2011}, and ballads of
heartbreak and survival, often performed by women whom queer audiences
took up as icons, such as Judy Garland \citep{dyer2004}.

Songs that explicitly address queer lives range from love songs to
protest. \citet{taylor2012} describes such music as a means of ``queer
world-making'': of creating spaces and shared feelings rather than
telling one story. Work on queer lives and feelings also questions the
expectation that queer lives be told as happy endings:
\citet{ahmed2010} analyses the demand for happiness that comes with
recognition, and \citet{love2007} the place of negative feelings in
queer history; the coming-out story of suffering, struggle and
liberation is one of the most conventional ``sexual stories''
\citep{plummer1995}.

\section{Data}
\label{sec:data}

\begin{figure*}[t]
\centering
\includegraphics[width=.85\textwidth]{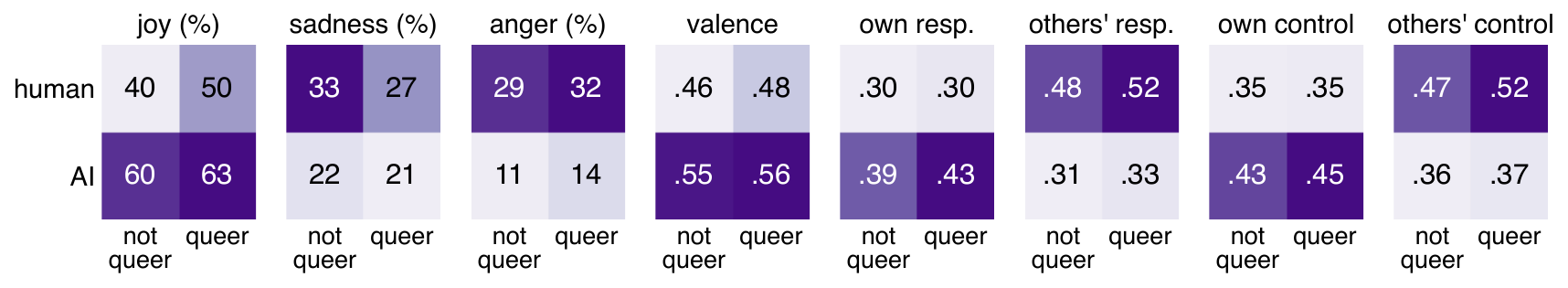}
\caption{Mean profiles of human and AI songs, not queer and queer
  (five shared genres): share of stanzas with joy, sadness and anger
  (\%), valence and the four appraisals of agency (0--1). Human ``not
  queer'': songs of the same genre and decade as the human queer
  songs. Darker cells: higher values within a panel.}
\label{fig:heat}
\end{figure*}

\begin{table*}[!t]
\centering\small\setlength{\tabcolsep}{4pt}
\begin{tabular}[t]{lrrrr}
\toprule
 & \multicolumn{1}{c}{\hq{}} & \multicolumn{1}{c}{\aq{}} & \multicolumn{1}{c}{\ag{}} & \multicolumn{1}{c}{\aq{}} \\
 & \multicolumn{1}{c}{$-$\hg{}} & \multicolumn{1}{c}{$-$\ag{}} & \multicolumn{1}{c}{$-$\hg{}} & \multicolumn{1}{c}{$-$\hq{}} \\
\midrule
\multicolumn{5}{l}{\textit{Emotion categories (pp)}} \\
\cmidrule(lr){1-5}
anger & +2.8$^*$ & +2.9$^*$ & $-$25.0$^*$ & $-$22.7$^*$ \\
disgust & +0.4\phantom{$^*$} & $-$0.8\phantom{$^*$} & $-$10.4$^*$ & $-$9.0$^*$ \\
fear & +0.2\phantom{$^*$} & $-$2.6$^*$ & $-$4.5$^*$ & $-$8.4$^*$ \\
joy & +9.9$^*$ & +3.1\phantom{$^*$} & +24.8$^*$ & +15.8$^*$ \\
sadness & $-$5.9$^*$ & $-$1.0\phantom{$^*$} & $-$10.5$^*$ & $-$9.3$^*$ \\
surprise & +0.8$^*$ & $-$0.4\phantom{$^*$} & $-$1.6$^*$ & $-$3.7$^*$ \\
shame/guilt & $\pm$0.0\phantom{$^*$} & $\pm$0.0$^*$ & $-$0.1$^*$ & $-$0.1\phantom{$^*$} \\
trust & +0.3\phantom{$^*$} & $-$4.7$^*$ & +6.7$^*$ & $-$0.9\phantom{$^*$} \\
no emotion & $-$5.1$^*$ & $-$2.6\phantom{$^*$} & +5.6\phantom{$^*$} & +12.3$^*$ \\
\addlinespace
\multicolumn{5}{l}{\textit{Affect (0--1)}} \\
\cmidrule(lr){1-5}
valence & +.024$^*$ & +.004\phantom{$^*$} & +.129$^*$ & +.098$^*$ \\
arousal & +.027$^*$ & +.012$^*$ & $-$.096$^*$ & $-$.122$^*$ \\
\bottomrule
\end{tabular}\hspace{2.5em}%
\begin{tabular}[t]{lrrrr}
\toprule
 & \multicolumn{1}{c}{\hq{}} & \multicolumn{1}{c}{\aq{}} & \multicolumn{1}{c}{\ag{}} & \multicolumn{1}{c}{\aq{}} \\
 & \multicolumn{1}{c}{$-$\hg{}} & \multicolumn{1}{c}{$-$\ag{}} & \multicolumn{1}{c}{$-$\hg{}} & \multicolumn{1}{c}{$-$\hq{}} \\
\midrule
\multicolumn{5}{l}{\textit{Appraisals (0--1)}} \\
\cmidrule(lr){1-5}
pleasantness & +.029$^*$ & +.021\phantom{$^*$} & +.139$^*$ & +.139$^*$ \\
unpleasantness & $-$.022$^*$ & $-$.002\phantom{$^*$} & $-$.161$^*$ & $-$.146$^*$ \\
suddenness & +.011\phantom{$^*$} & $-$.003\phantom{$^*$} & $-$.115$^*$ & $-$.139$^*$ \\
familiarity & $-$.011$^*$ & +.014$^*$ & +.014$^*$ & +.054$^*$ \\
goal relevance & +.009$^*$ & +.010$^*$ & $-$.046$^*$ & $-$.053$^*$ \\
goal conducive. & +.017$^*$ & +.009\phantom{$^*$} & +.137$^*$ & +.139$^*$ \\
own resp. & +.006\phantom{$^*$} & +.036$^*$ & +.110$^*$ & +.154$^*$ \\
others' resp. & +.048$^*$ & +.013\phantom{$^*$} & $-$.180$^*$ & $-$.215$^*$ \\
situat. resp. & $-$.013$^*$ & $-$.017\phantom{$^*$} & $-$.027$^*$ & $-$.056$^*$ \\
own control & +.002\phantom{$^*$} & +.021\phantom{$^*$} & +.100$^*$ & +.134$^*$ \\
others' control & +.042$^*$ & +.007\phantom{$^*$} & $-$.126$^*$ & $-$.155$^*$ \\
situat. control & $-$.021$^*$ & $-$.016\phantom{$^*$} & $-$.012\phantom{$^*$} & $-$.038$^*$ \\
attention & +.002\phantom{$^*$} & $-$.007\phantom{$^*$} & +.056$^*$ & +.037$^*$ \\
effort & $-$.011\phantom{$^*$} & $-$.002\phantom{$^*$} & $-$.086$^*$ & $-$.101$^*$ \\
urgency & +.011$^*$ & +.004\phantom{$^*$} & $-$.090$^*$ & $-$.098$^*$ \\
certainty & $-$.005\phantom{$^*$} & +.027$^*$ & +.002\phantom{$^*$} & +.076$^*$ \\
\bottomrule
\end{tabular}
 \caption{Differences between the groups for all outputs of the model
  (acronyms, see Figure~\ref{fig:sketch}; in \mbox{\hq{}$-$\hg}, \hg{}
  has the genre and decade of \hq). Categories: share of stanzas in
  percentage points (pp); valence, arousal and appraisals on a 0--1
  scale.}
\label{tab:contrasts}
\end{table*}

\subsection{Human Songs}
We use 3.2 million English songs from Genius in the cleaned version of
\citet{dickinson2026genius},\footnote{\url{https://huggingface.co/datasets/theelderemo/genius-lyrics-cleaned}}
with title, artist, year and one of five genre tags (pop, rap, rock,
R\&B, country). We collect human songs associated with queer
communities from two Wikipedia sources: the ``List of gay anthems''
\citep{wiki-anthems}, songs the community took up (355 songs, 219
found in Genius), and the ``LGBTQ-related songs'' category
\citep{wiki-category}, songs about queer topics or by queer artists
(561 songs, 344 found). We match title and artist, allow at most five
years between the release years, and keep one Genius version per song:
490 human queer songs by 302 artists. We consider all other Genius
songs as human songs that are not queer.

\subsection{AI Songs}
The Sounds Queer data set
\citep{weber2026soundsqueer} contains songs generated by Suno and
Mureka for the prompt ``Make a \textit{genre} song about a
\textit{orientation} \textit{gender} \textit{man/woman/person}'' with
ten genres and optional orientation and gender words, four generations
per prompt and service (5,963 songs after removing duplicate
lyrics). A song is \emph{AI queer} if its prompt has a queer
orientation (e.g., lesbian, bisexual, asexual) or a trans, intersex or
non-binary gender word (5,366 songs, 2,160 prompts), and \emph{AI not
  queer} otherwise (``straight'', ``cis'' or no word; 597 songs, 240
prompts). Five genres occur in both sources: pop, rock, hip hop
(Genius: rap), country and metal.\footnote{Genius has no metal tag; we
  count rock-tagged songs by bands on Wikipedia's lists of heavy metal
  bands \citep{wiki-metal}.}\textsuperscript{,}\footnote{Genius lyrics
  are copyrighted; we analyse them computationally and quote only song
  titles. Sounds Queer is published under CC BY-NC 4.0.}

\section{Method}
\subsection{Overview}
Figure~\ref{fig:sketch} shows our approach. We assign every song to
one of the four groups of the $2\times2$ design
(Section~\ref{sec:data}), split it into stanzas and let an emotion
model predict emotion categories, valence and arousal and appraisals
for every stanza (\ref{sec:model}). From the stanza predictions we
build two representations of a song: its profile, the average over all
stanzas, and its course, the predictions by position in the song. The
profiles are the basis for the comparison of the groups
(\ref{sec:compare}; Q1--Q3 in Section~\ref{sec:results}), the course
is the basis for the analysis of narratives (\ref{sec:course}; Q4).

\subsection{Emotion Model}
\label{sec:model}
We split each song into stanzas at blank lines and section headers
(e.g., \textit{[Chorus]}), count a repeated stanza (such as a chorus)
once, and predict each stanza with the general model of the BamNLP
Emotion Hub \citep{klinger2027hub}. It is an XLM-RoBERTa encoder
\citep{conneau2020xlmr} fine-tuned on the English and German emotion
corpora with human annotations collected in the hub. For a text it
predicts nine emotion categories (anger, disgust, fear, joy, sadness,
surprise, shame/guilt, trust, no emotion; several can hold), valence
and arousal, and 16 appraisal dimensions. An event gate decides
whether a text describes an event; appraisals are only interpreted for
such texts. The model was not trained on lyrics; on corpora it has not
seen, it keeps about 70\% of the performance of a model trained on
that corpus.

\begin{table*}
\centering\small\setlength{\tabcolsep}{3.5pt}
\begin{tabular}{lrrrrrrrrrrrrr}
\toprule
 & \multicolumn{1}{c}{\hq} & \multicolumn{3}{c}{joy} & \multicolumn{3}{c}{valence} & \multicolumn{3}{c}{others' resp.} & \multicolumn{3}{c}{own control} \\
\cmidrule(lr){3-5}\cmidrule(lr){6-8}\cmidrule(lr){9-11}\cmidrule(lr){12-14}
genre & \multicolumn{1}{c}{songs} & \multicolumn{1}{c}{\hq{}} & \multicolumn{1}{c}{\aq{}} & \multicolumn{1}{c}{\ag{}} & \multicolumn{1}{c}{\hq{}} & \multicolumn{1}{c}{\aq{}} & \multicolumn{1}{c}{\ag{}} & \multicolumn{1}{c}{\hq{}} & \multicolumn{1}{c}{\aq{}} & \multicolumn{1}{c}{\ag{}} & \multicolumn{1}{c}{\hq{}} & \multicolumn{1}{c}{\aq{}} & \multicolumn{1}{c}{\ag{}} \\
 & & \multicolumn{1}{c}{$-$\hg{}} & \multicolumn{1}{c}{$-$\ag{}} & \multicolumn{1}{c}{$-$\hg{}} & \multicolumn{1}{c}{$-$\hg{}} & \multicolumn{1}{c}{$-$\ag{}} & \multicolumn{1}{c}{$-$\hg{}} & \multicolumn{1}{c}{$-$\hg{}} & \multicolumn{1}{c}{$-$\ag{}} & \multicolumn{1}{c}{$-$\hg{}} & \multicolumn{1}{c}{$-$\hg{}} & \multicolumn{1}{c}{$-$\ag{}} & \multicolumn{1}{c}{$-$\hg{}} \\
\midrule
pop & 311 & +12$^*$ & +1\phantom{$^*$} & +30$^*$ & +.03$^*$ & $-$.01\phantom{$^*$} & +.15$^*$ & +.05$^*$ & $-$.02\phantom{$^*$} & $-$.14$^*$ & +.01\phantom{$^*$} & +.03\phantom{$^*$} & +.12$^*$ \\
rock & 94 & +10$^*$ & +6\phantom{$^*$} & +13\phantom{$^*$} & +.02\phantom{$^*$} & +.02\phantom{$^*$} & +.07$^*$ & +.06$^*$ & +.01\phantom{$^*$} & $-$.13$^*$ & +.01\phantom{$^*$} & +.04\phantom{$^*$} & +.04\phantom{$^*$} \\
hip hop & 25 & +6\phantom{$^*$} & +9\phantom{$^*$} & +27$^*$ & +.02\phantom{$^*$} & $-$.01\phantom{$^*$} & +.21$^*$ & +.03\phantom{$^*$} & $\pm$.00\phantom{$^*$} & $-$.32$^*$ & $-$.01\phantom{$^*$} & $\pm$.00\phantom{$^*$} & +.25$^*$ \\
country & 20 & +2\phantom{$^*$} & $-$1\phantom{$^*$} & +27$^*$ & $-$.01\phantom{$^*$} & $\pm$.00\phantom{$^*$} & +.12$^*$ & +.05\phantom{$^*$} & +.04$^*$ & $-$.20$^*$ & $\pm$.00\phantom{$^*$} & $\pm$.00\phantom{$^*$} & +.06$^*$ \\
metal & 8 & +11\phantom{$^*$} & +1\phantom{$^*$} & +27$^*$ & $-$.03\phantom{$^*$} & +.02\phantom{$^*$} & +.09$^*$ & +.20$^*$ & +.04\phantom{$^*$} & $-$.11$^*$ & $-$.05\phantom{$^*$} & +.03\phantom{$^*$} & +.02\phantom{$^*$} \\
\midrule
all five & 490 & +10$^*$ & +3\phantom{$^*$} & +25$^*$ & +.02$^*$ & $\pm$.00\phantom{$^*$} & +.13$^*$ & +.05$^*$ & +.01\phantom{$^*$} & $-$.18$^*$ & $\pm$.00\phantom{$^*$} & +.02\phantom{$^*$} & +.10$^*$ \\
\bottomrule
\end{tabular}
 \caption{Differences per genre (acronyms, see Figure~\ref{fig:sketch}; in \mbox{\hq{}$-$\hg}, \hg{} has the genre and
decade of \hq). Joy in pp. \hq{} songs: number of songs in the genre (``all five'' incl.\ 32 R\&B songs).}
\label{tab:genres}
\end{table*}

\begin{table*}
\centering\small\setlength{\tabcolsep}{4pt}
\begin{tabular}{lrrrrrrrrr}
\toprule
 & \multicolumn{1}{c}{\hq} & \multicolumn{4}{c}{anger} & \multicolumn{4}{c}{disgust} \\
\cmidrule(lr){3-6}\cmidrule(lr){7-10}
genre & \multicolumn{1}{c}{songs} & \multicolumn{1}{c}{\hq{}$-$\hg{}} & \multicolumn{1}{c}{\aq{}$-$\ag{}} & \multicolumn{1}{c}{\ag{}$-$\hg{}} & \multicolumn{1}{c}{\aq{}$-$\hq{}} & \multicolumn{1}{c}{\hq{}$-$\hg{}} & \multicolumn{1}{c}{\aq{}$-$\ag{}} & \multicolumn{1}{c}{\ag{}$-$\hg{}} & \multicolumn{1}{c}{\aq{}$-$\hq{}} \\
\midrule
pop & 311 & +2.2\phantom{$^*$} & +3.1\phantom{$^*$} & $-$24.1$^*$ & $-$22.7$^*$ & $-$1.3\phantom{$^*$} & $-$1.2\phantom{$^*$} & $-$7.2$^*$ & $-$7.0$^*$ \\
rock & 94 & +3.2\phantom{$^*$} & +2.8\phantom{$^*$} & $-$15.2$^*$ & $-$13.2$^*$ & +2.2\phantom{$^*$} & +0.7\phantom{$^*$} & $-$10.5$^*$ & $-$10.3$^*$ \\
hip hop & 25 & +6.9\phantom{$^*$} & +3.6\phantom{$^*$} & $-$55.7$^*$ & $-$59.1$^*$ & +0.9\phantom{$^*$} & $-$0.9\phantom{$^*$} & $-$20.5$^*$ & $-$22.2$^*$ \\
country & 20 & +2.6\phantom{$^*$} & +1.7\phantom{$^*$} & $-$19.5$^*$ & $-$19.3$^*$ & +4.6\phantom{$^*$} & $-$1.8$^*$ & $-$4.2$^*$ & $-$10.3$^*$ \\
metal & 8 & +19.7$^*$ & +3.5\phantom{$^*$} & $-$10.6$^*$ & $-$27.0$^*$ & +17.3\phantom{$^*$} & $-$0.5\phantom{$^*$} & $-$9.7$^*$ & $-$27.7$^*$ \\
\midrule
all five & 490 & +2.8$^*$ & +2.9$^*$ & $-$25.0$^*$ & $-$22.7$^*$ & +0.4\phantom{$^*$} & $-$0.8\phantom{$^*$} & $-$10.4$^*$ & $-$9.0$^*$ \\
\bottomrule
\end{tabular}
 \caption{Anger and disgust per genre (share of stanzas, pp; acronyms, see Figure~\ref{fig:sketch}; in \mbox{\hq{}$-$\hg},
\hg{} has the genre and decade of \hq).}
\label{tab:negative}
\end{table*}

\subsection{Song Profiles and Group Comparisons}
\label{sec:compare}
A song's profile is the share of its stanzas with each emotion
category and the mean valence, arousal and appraisals over its
stanzas. We compare human queer songs with human songs of the same
genre and decade, AI queer with AI not-queer songs within each genre,
and AI with human songs within each genre (the five shared genres,
weighted equally). We write \hg{} and \hq{} for human songs that are
not queer and queer, and \ag{} and \aq{} for AI songs that are not
queer and queer. A difference counts as significant (marked $^*$ in
the tables) if its 95\% bootstrap interval excludes 0; the bootstrap
resamples artists (human queer songs) and prompts (AI songs), so that
songs of one artist or one prompt do not count as independent.

\subsection{Course of a Song}
\label{sec:course}
To capture a narrative, we give each stanza its relative position in
the song, from 0 (first stanza) to 1 (last stanza). For a detailed
picture, we average the predictions within each fifth of the song
(positions 0--0.2, \ldots, 0.8--1) and then over the songs of a
group. For a single number per song, its \emph{arc}, we subtract the
mean of the first half (positions below 0.5) from the mean of the
second half (above 0.5). The fifths show the shape of the course, the
arc allows a test of whether a group's songs end more positively than
they begin.

\section{Results}
\label{sec:results}

Our research question asks how human-written and AI-generated songs on
queer topics differ in the emotions and appraisals they describe. We
answer it in four steps. Q1 and Q2 establish the two baselines of the
$2\times2$ design: how human queer songs differ from other human
songs, and how AI songs differ from human songs in general. Q3 asks
whether the AI reproduces the human difference. Q4 turns from averages
to the form of the songs: their course. Figure~\ref{fig:heat} shows
the most frequent emotions, valence and the appraisals of agency,
i.e., whether people act or are acted upon.

\paragraph{Q1: Are human queer songs emotionally different from other
  human songs?} Column \mbox{\hq{}$-$\hg} of Table~\ref{tab:contrasts}
and the human row of Figure~\ref{fig:heat} answer this. Human queer
songs have joy in 10 percentage points more of their stanzas than
songs of the same genre and decade and sadness in 6 points fewer; they
are more aroused and more often appraise events as caused and
controlled by other people. The difference holds in pop and rock,
where most human queer songs are (Table~\ref{tab:genres},
\mbox{\hq{}$-$\hg}). Human queer songs are more joyful, but their joy
addresses others.

\begin{figure*}[t]
\centering
\includegraphics[width=.86\textwidth]{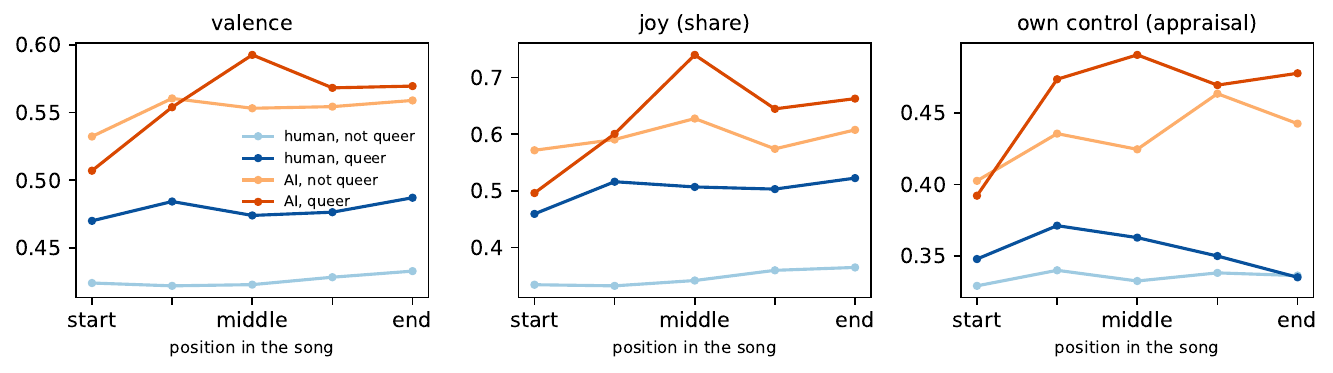}
\caption{Course of a song: valence, share of stanzas with joy and own control in each fifth of the song (shared genres; Section~\ref{sec:course}). The arc in the text compares the second with the first half.}
\label{fig:traj}
\end{figure*}

\paragraph{Q2: Are AI songs more positive than human songs in
  general?} Column \mbox{\ag{}$-$\hg} of Table~\ref{tab:contrasts} and
the left column of each heatmap compare AI and human songs without a
queer topic. AI songs have joy in 25 points more of their stanzas,
anger in 25 points fewer, and they appraise events as more pleasant
and more under their own control, less as caused by
others. Table~\ref{tab:genres} (\mbox{\ag{}$-$\hg}) shows this in
every genre, including the genres whose human lyrics are least
joyful. It holds for human songs from 2010 onwards and for prompts
that name no identity at all (joy $+24$ points). AI songs are positive
regardless of their topic.

\paragraph{Q3: Does the AI reproduce the difference between queer and
  other songs?} Column \mbox{\aq{}$-$\ag} of Table~\ref{tab:contrasts}
shows that a queer identity in the prompt raises joy by only 3 points
(not significant), against 10 points for human songs; the difference
between the two gaps is $-6.8$ points. In no genre does a queer prompt
make AI songs reliably more joyful or more pleasant
(Table~\ref{tab:genres}, \mbox{\aq{}$-$\ag}). The AI does not learn
that queer songs are happier; it makes every song happy. Split by the
kind of identity in the prompt, the picture is mixed: over all stanzas
of the ten genres, the gender words trans, intersex and non-binary
yield 5.3 points less joy than the gender word ``cis'', while queer
orientations (e.g., lesbian, bisexual) yield 5.4 points more than the
orientation ``straight''.

\textit{Genre-typical negative emotions.} Human queer songs have
more anger than human songs of the same genre and decade in all five
genres, least in pop (+2.2 points) and then in country, rock and hip
hop (+2.6 to +6.9), most in metal (+19.7; Table~\ref{tab:negative},
\mbox{\hq{}$-$\hg}). AI songs have much less anger than human songs in
every genre, but not in proportion: in metal, AI songs without a queer
topic have 10.6 points less anger than human songs, AI queer songs
27.0 points less than human queer songs (\mbox{\ag{}$-$\hg} and
\mbox{\aq{}$-$\hq}). Disgust in country shows the same pattern ($-4.2$
and $-10.3$ points), and here a queer prompt also lowers disgust
within the AI songs ($-1.8$ points, the only significant difference
\mbox{\aq{}$-$\ag} in the table). A ``normal'' AI metal song may thus
be as angry as its genre expects, but a queer one is no angrier,
although human queer metal songs are: the AI drops the genre-typical
negative emotions of queer songs and tells them, also in metal, as
stories of self-realisation and acceptance. One
explanation is that the generators avoid negative emotions in the
context of queerness.

\paragraph{Q4: Do AI queer songs follow a narrative script?}
Figure~\ref{fig:traj} shows the course of a song from its first to its
last fifth. In AI queer songs, joy rises by 10 points from the first
to the second half and sadness falls by 8 points; in human songs,
queer or not, and in AI songs without a queer topic, the changes are
small. 43\% of AI queer songs rise in valence by more than 0.05,
against 30--34\% in the other groups.  Compared with prompts that name
no identity, the rise in joy is 16 points larger.  An analysis of
examples shows that AI queer songs with the largest rise in valence
share the plot that external societal forces impose labels upon the
narrator of the lyrics, the narrator refuses the classification and
the song ends in the narrator reclaiming their freedom.  The human
queer songs follow no such plot.

\section{Discussion}

Our design separates two explanations for positive AI songs about
queer people. The positivity is not a learned property of queer songs:
human queer songs are happier than other human songs, but the AI's
queer songs are hardly happier than its other songs, and all AI songs
are far happier than human songs in every genre. What marks the AI's
queer songs is form: one narrative script, from being labelled and
confined by others to self-assertion and freedom.  This result
corresponds to the positive stereotype that \citet{cheng2023marked}
found in generated personas. The generators vary surface features such
as genre and imagery, but hardly the emotion structure.

\section{Conclusion}

Human songs associated with queer communities are more joyful than
other songs; AI-generated songs are far more joyful than human songs,
whether or not they are about queer people. The AI's queer songs
differ from its other songs in their plot: from sadness and
confinement to joy and self-determination. Positive stereotyping can
thus be measured as a narrative script.

One interpretation of these results is \emph{emotion flattening}: the
AI does not reproduce what makes human queer songs distinct, but
removes these differences. Human queer songs have more joy than other
human songs; in AI songs, this difference disappears, because every AI
song is joyful and has little anger or disgust.  Over all genres, the
small rise in anger is reproduced, but not where the genre calls for
anger: in metal, human queer songs are far angrier than other songs,
AI queer songs are not. Where the AI's queer songs differ from its
other songs, they differ in their script, not in the emotions that set
human queer songs apart.

\section*{Disclosure of the Use of Generative AI}
The Sounds Queer songs that we analyse were generated by Suno and
Mureka \citep{weber2026soundsqueer}. The analysis code and a first
draft of this paper were produced with the help of Claude (Anthropic)
on the authors' instructions; the authors designed the study, checked
the analyses and are responsible for the text.

\end{document}